\documentclass[sigconf,authorversion=true]{acmart}

\usepackage{amsmath,amssymb,amsfonts}
\usepackage{algorithmic}
\usepackage{graphicx}
\usepackage{xspace}
\usepackage{textcomp}
\usepackage{colortbl}
\usepackage{multirow}
\usepackage{array}
\usepackage{makecell}
\usepackage{subcaption}
\usepackage{tikz}
\usepackage{xcolor}
    
\newcommand*\circled[1]{\tikz[baseline=(char.base)]{%
    \node[shape=circle,fill=black,text=white,inner sep=1pt,font=\bfseries] (char) {#1};}}

\newcommand{\tabref}[1]{Table~\ref{#1}}

\newcommand\tta{\textsc{TTA}\xspace}

\newcommand{\etal}{\emph{et~al.}\xspace}

\newcommand{\eg}{\emph{e.g.},\xspace}
\newcommand{\ie}{\emph{i.e.},\xspace}

\newcommand{\dnn}{DNN\xspace}
\newcommand{\mlp}{MLP\xspace}

\newcommand{\rapsd}{RAPSD\xspace}
\newcommand{\dct}{DCT\xspace}
\newcommand{\dwt}{DWT\xspace}
\newcommand{\tent}{TENT\xspace}
\newcommand{\cdag}{DAG\xspace}
\newcommand{\nfjf}{NFJF\xspace}
\newcommand{\isp}{ISP\xspace}



\author{Francesco Corti}
\authornote{Equal contribution}
\affiliation{%
  \institution{Graz University of Technology}
  \city{Graz}
  \country{Austria}
}
\email{francesco.corti@tugraz.at}
\author{Moritz Becker}
\authornotemark[1]
\affiliation{%
  \institution{Graz University of Technology}
  \city{Graz}
  \country{Austria}
}
\email{m.becker@student.tugraz.at}
\author{Olga Saukh}
\affiliation{%
  \institution{Graz University of Technology \\ Complexity Science Hub}
  \city{Graz, Vienna}
  \country{Austria}
}
\email{saukh@tugraz.at}

\copyrightyear{2026}
\acmYear{2026}
\setcopyright{cc}
\setcctype{by}
\acmConference[IOT 2026]{The 16th International Conference on the Internet of Things}{November 17--20, 2026}{Newcastle Upon Tyne, England, United Kingdom}
\acmBooktitle{The 16th International Conference on the Internet of Things (IOT 2026), November 17--20, 2026, Newcastle Upon Tyne, England, United Kingdom}
\acmDOI{10.1145/3860552.3860562}
\acmISBN{979-8-4007-3095-5/2026/11}

\begin{document}

\title{No Frames, Just Frequencies: Bandwidth-Efficient \\ Deployment-Time Adaptation for Smart Glasses}

\begin{abstract}
Deep neural networks deployed on smart glasses must adapt to changes in the distribution of incoming data while adhering to strict privacy and device constraints. We present \emph{No Frames, Just Frequencies}, a bandwidth-efficient source-assisted adaptation framework that characterizes distribution shifts using compact frequency descriptors computed on-device. Rather than transmitting camera frames, the glasses compute and send frequency descriptors to a host device that contains a replica of the model. The host device maps the received frequency descriptor to a distribution shift operator that synthesizes out-of-distribution data and adapts a model replica using a pre-defined adaptation protocol. We evaluate the framework on mixtures of ImageNet-C corruptions using convolution-based (ConvNeXtV2-Femto), self-attention-based (DeiT-Tiny), and hybrid (TinyViT-5M) vision models, on person classification (Visual Wake Words), and on object detection (MS-COCO). Our results reveal that adaptation driven solely by the transmitted descriptors recovers $93-95\%$ of the supervised oracle adaptation gain on all three architectures, $82\%$ on Visual Wake Words and $94$--$98\%$ on MS-COCO.  Notably, adaptation quality degrades gracefully as the descriptor resolution decreases from $224 $ to $28$ pixels, consistently across architectures, enabling bandwidth reductions from $213\times$ to $784\times$ across the frequency descriptors.
\end{abstract}

\maketitle

\section{Introduction}
Deep neural network (\dnn) models deployed at the edge often lack robustness when the input data~\cite{hendrycks2019benchmarking} drift from the training data,  degrading the model's prediction quality~\cite{saukh2023representing}. Practical deployments must additionally cope with dynamic resource constraints (\eg fluctuating energy budgets and latency requirements), which may require rapid, runtime reconfiguration of the deployed model~\cite{corti2025reds}. 

Smart glasses represent an especially demanding edge setting: they must continuously process video streams within tight latency and power-efficiency budgets while operating in close physical proximity to the user~\cite{karakostas2024realtimeAR}. The motivation for on-device intelligence is therefore dual: local inference minimizes dependence on network connectivity~\cite{zhou2019edge}, and processing video data on-device reduces the exposure of raw visual content, which on smart glasses may contain sensitive information (\eg faces, screens, locations) that should not leave the device.

Privacy is not the only constraint: smart glasses also operate under severe hardware resource limitations. While edge-designed convolutional architectures such as MobileNets~\cite{howard2017mobilenets} and EfficientNet~\cite{tan2019efficientnet} have made inference practical on edge devices, adaptation remains the bottleneck for both convolution- and attention-based models~\cite{lin2023tiny}, since gradient-based methods must store intermediate activations and optimizer states, substantially increasing the memory footprint beyond inference alone~\cite{slamanig2025, xu2022mandheling}. As a wearable camera moves through heterogeneous environments, the visual stream experiences rapid changes in illumination and motion blur, producing non-stationary shifts over time~\cite{quinonero2008dataset,yao2022wild} that degrade the predictive quality of the deployed model, thus adaptation must operate continuously during deployment.

Test-Time Adaptation (\tta) adapts pre-trained models to changes in data distribution at test time by using each data point once for post-inference adaptation~\cite{sun2020test}. Existing \tta methods expose two methodological gaps for resource-constrained edge devices, such as smart glasses. First, popular approaches such as entropy minimization~\cite{wang2020tent} may fail under certain shifts and amplify the model’s prediction errors without safeguards~\cite{press2024entropy}. Second, they assume mini-batches large enough (or an auxiliary cache) to estimate accurate target statistics. Under realistic on-device constraints, where only a few samples may be available, the model’s performance can degrade sharply and may even collapse for online adaptation batches due to unreliable statistics estimation or noisy gradient adaptation~\cite{su2024unraveling, niu2023towards}. Prior work mitigates these methodological gaps by emulating larger effective batches with gradient accumulation~\cite{lin2023improving} or by constructing a virtual batch via test-time augmentations~\cite{zhang2022memo}, which conflicts with strict memory and privacy restrictions when storing or replaying visual samples on smart glasses.

\begin{figure*}[t]
    \centering
    \includegraphics[width=0.85\textwidth]{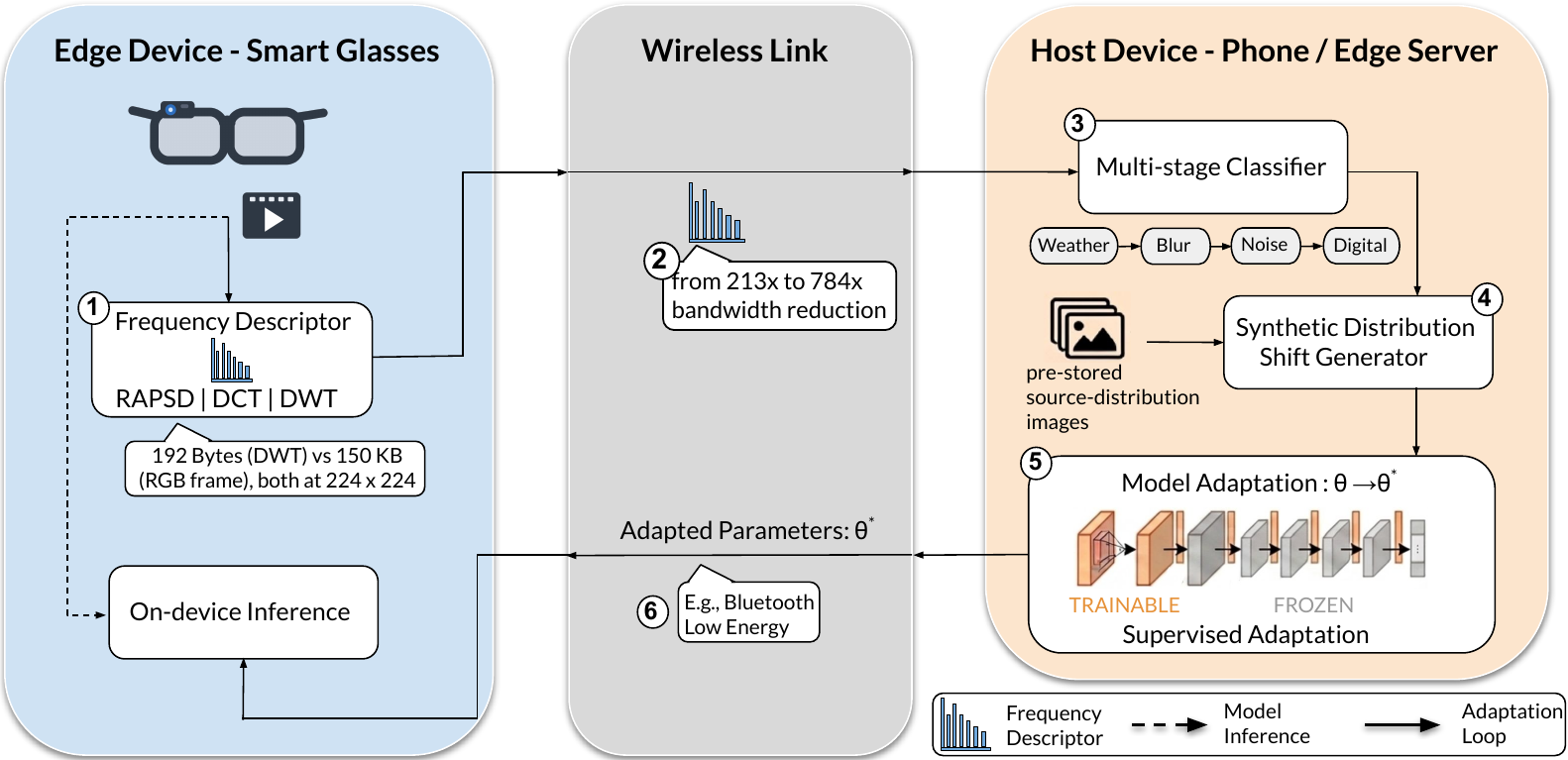}
    \caption{\textbf{System overview of the proposed bandwidth-efficient \tta framework.} An edge device (\ie Smart Glasses) computes compact frequency descriptors (\rapsd, DCT, or DWT statistics) from incoming camera frames on-device. Only these descriptors are transmitted to the host device, keeping the per-frame uplink within a low-bandwidth budget instead of streaming raw frames. The host device maps the received descriptor to a synthetic corruption distribution through a sequential pipeline of training corruptions. The host device generates synthetic corrupted data batches from locally stored in-distribution images, and adapts a model replica using a pre-defined adaptation protocol~\cite{lee2023surgical, wang2020tent}. Finally, the updated parameter subset is transmitted back to the edge device.} 
    \label{fig:system_overview}
\end{figure*}

Our approach to \emph{bandwidth-efficient} deployment-time adaptation satisfies three requirements: (\emph{i}) it avoids streaming camera frames over the bandwidth-limited uplink by relying on compact, privacy-enhancing distribution-shift frequency descriptors computed on-device; (\emph{ii}) it offloads the gradient adaptation step to a nearby host device to respect the glasses' tight compute and memory budgets, restricting the adaptation to a small parameter subset; (\emph{iii}) it couples adaptation with an explicit characterization of the distribution shift affecting the model's predictions under non-stationary streams. Concretely, the edge device computes a compact frequency descriptor for each frame and transmits it to a host device (Fig.~\ref{fig:system_overview}). These descriptors capture only the spectral magnitude, discarding the phase information and aggregating the spectrum into as few as 14 floating-point values, so that the per-frame uplink fits within a low-bandwidth wireless link budget. The host device maps the received descriptor to a four-stage causal corruption directed acyclic graph (\cdag), synthesizes adaptation batches from pre-stored in-distribution data, adapts a model replica on a compact subspace via supervised cross-entropy, and transmits the adapted parameters back to the smart glasses, thereby closing the adaptation loop without transmitting camera frames; we term this setting \emph{source-assisted adaptation}. Our contributions are:
\begin{itemize}
    \item We propose \emph{No Frames, Just Frequencies} (\nfjf), a bandwidth-efficient source-assisted adaptation framework for smart glasses that characterizes distribution shifts using compact on-device frequency descriptors and offloads adaptation to a paired host.
    \item We design a multi-stage corruption classifier that maps a frequency descriptor to a four-stage causal corruption directed acyclic graph~(Weather $\rightarrow$ Blur $\rightarrow$ Noise $\rightarrow$ Digital), paired with a synthetic data-generation pipeline that reconstructs the detected distribution shift from labeled in-distribution data samples.
    \item We conduct a systematic evaluation across three architecture families (convolution, self-attention, and hybrid), three frequency descriptors (\rapsd, DCT, DWT), and four input resolutions, against $63$  mixtures of corruptions.  
    \item We extend the framework to person classification and object detection, analyze the adaptation accuracy of the frequency descriptor against model-based distribution shift descriptors, and measure the end-to-end latency of the adaptation loop on smart-glasses class hardware.   
\end{itemize}

\section{Related Work}
\label{sec:related}

\textbf{Test-Time Adaptation} (\tta) methods adapt pre-trained models to distribution shifts at deployment without access to the source training data or target labels~\cite{sun2020test}. Test-time Adaptation by Entropy Minimization~(\tent)~\cite{wang2020tent} minimizes the entropy of model predictions by re-estimating normalization statistics from the test batch and updating the affine parameters of normalization layers. SAR~\cite{niu2023towards} mitigates small-batch instability through sharpness-aware minimization and reliable sample filtering, and SPA~\cite{niu2025self} enforces a consistency objective between clean and augmented inputs of Vision Transformers. Unlike these methods, which adapt on the unlabeled test stream itself, our framework keeps the test frames on the glasses and offloads gradient computation to a host that adapts on labeled, in-distribution images; it operates in a \emph{source-assisted} regime rather than under the source-free, label-free assumption of \tta, in exchange for a bandwidth-efficient wireless transmission.

\textbf{Adaptation Beyond \tta.}
Unsupervised domain adaptation assumes joint access to labeled source and unlabeled target images~\cite{ganin2015unsupervised}, whereas source-free domain adaptation discards the source data and adapts on unlabeled target samples alone~\cite{liang2020we}; continual \tta extends this setting to non-stationary target streams~\cite{ liang2025comprehensive}. Predictive domain adaptation infers the parameters of the target model from domain metadata, without any target data~\cite{mancini2019adagraph} and zero-shot domain adaptation generates a model for an unseen domain from a semantic domain descriptor~\cite{yang2014unified}. These approaches either observe target samples or specify the expected target conditions prior to deployment; \nfjf instead estimates the shift online from the transmitted frequency descriptors and synthesizes the corresponding labeled adaptation data on the host. 

\textbf{Domain Adaptation via Frequency Manipulation.} Prior work~\cite{yang2020fda} shows that swapping the low-frequency amplitude components between source and target images supports semantic segmentation domain adaptation: the low-frequency amplitude tends to capture the style and domain-specific characteristics, while the phase encodes semantic content~\cite{xu2021fourier}. Subsequent work on the frequency bias of deep neural networks~\cite{yin2019fourier} shows that different corruption families concentrate their perturbation energy in distinct bands of the spatial-frequency spectrum, a property we exploit to classify corruptions from descriptors. In contrast to prior work, which manipulates frequencies for offline adaptation, we use frequency descriptors as a bandwidth-efficient \emph{communication primitive} between an edge device and a host, and as the input to the corruption classifier that drives the synthesis of the labeled adaptation data. 

\textbf{Privacy-Enhancing Edge Intelligence.} Edge computing minimizes data transmission to reduce latency and preserve privacy~\cite{shi2016edge,zhou2019edge}; federated learning~\cite{mcmahan2017communication} exchanges gradients rather than data, which requires on-device gradient computation, often not available on constrained edge devices such as smart glasses. Our framework transmits only frequency descriptors, whose phase is discarded, providing bandwidth savings over camera frame streaming and a privacy-enhancing alternative to gradient-sharing methods. 

\section{Background}
\label{sec:background}

We consider a pair of smart glasses equipped with a camera sensor running a pre-trained vision model $f_\theta$, with parameters $\theta$. At each timestep $t$, the device receives a visual frame $x_t \in \mathbb{R}^{H \times W \times 3}$ and produces a prediction $\hat{y}_t = f_\theta(x_t)$. Our proposed system (Fig.~\ref{fig:system_overview}) adapts $f_\theta$ to the predicted environment shift without exposing visual content in six steps: 

\circled{1}~The glasses compute a compact frequency descriptor: $\mathbf{d}_t = \Phi(x_t) \in \mathbb{R}^D$ for each incoming frame $x_t$, where $\Phi$ is one of the descriptor functions of Sec.~\ref{sec:method}. The descriptor $\mathbf{d}_t$ comprises as few as 14 floats at $28 \times 28$ input resolution; at the deployed model's native $224 \times 224$ input, the DWT descriptor (192 B) replaces the 150,528 B raw RGB frame it summarizes, reducing the transmitted data by 784$\times$; across the three descriptors at the native input, the per-frame reduction ranges from 213$\times$ (\dct, $708$\,B) to $784\times$ (\dwt, $192$\,B). 

\circled{2}~The descriptor $\mathbf{d}_t$ is transmitted to a host device (\eg a paired phone) over a bandwidth-limited wireless link. Because the phase is discarded and the magnitude pooled, only a lossy summary of the frame leaves the device. 

\circled{3}~The host runs a multi-stage corruption classifier $g_\phi$, pre-trained on descriptors of mixtures of corruptions modeled as Directed Acyclic Graphs (\cdag{}s), which maps, for each stage $s$ $\mathbf{d}_t$ to a discrete distribution $P(c_s, \ell_s \mid s, \mathbf{d}_t)$ over the stage's corruption-severity pairs, with $\ell_s = 0$ as the inactive stage. 

\circled{4}~ A synthetic distribution shift generator constructs labeled adaptation batches by composing the predicted corruptions in the camera's image signal processing (\isp)  order over a set of in-distribution pre-stored data (Sec. \ref{sec:classifier}). 

\begin{figure*}[t]
\centering
\begin{subfigure}[t]{0.33\textwidth}
  \centering
  \includegraphics[width=\textwidth]{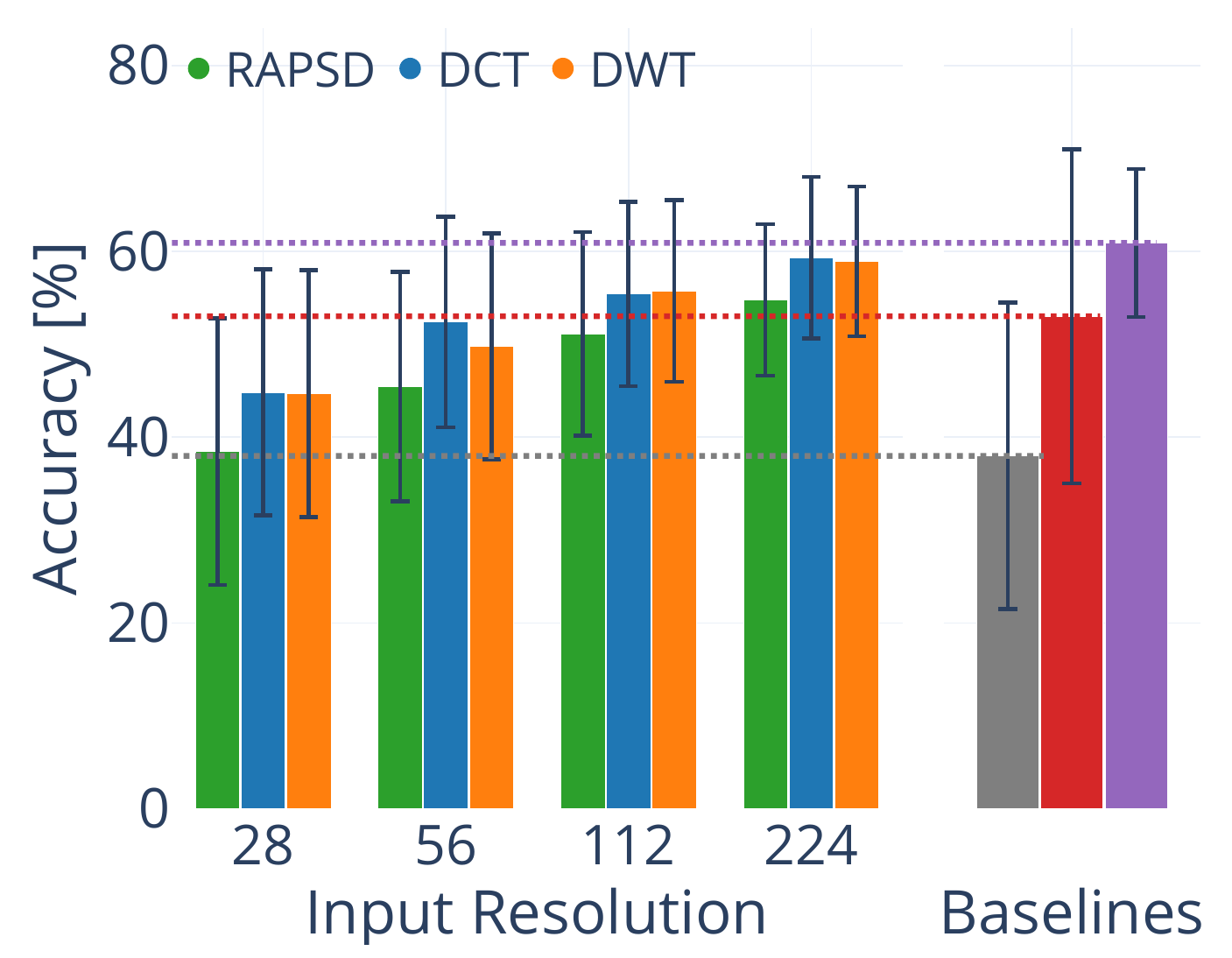}
  \caption{ConvNeXtV2-Femto.}
  \label{fig:bars_convnext}
\end{subfigure}\hfill
\begin{subfigure}[t]{0.33\textwidth}
  \centering
  \includegraphics[width=\textwidth]{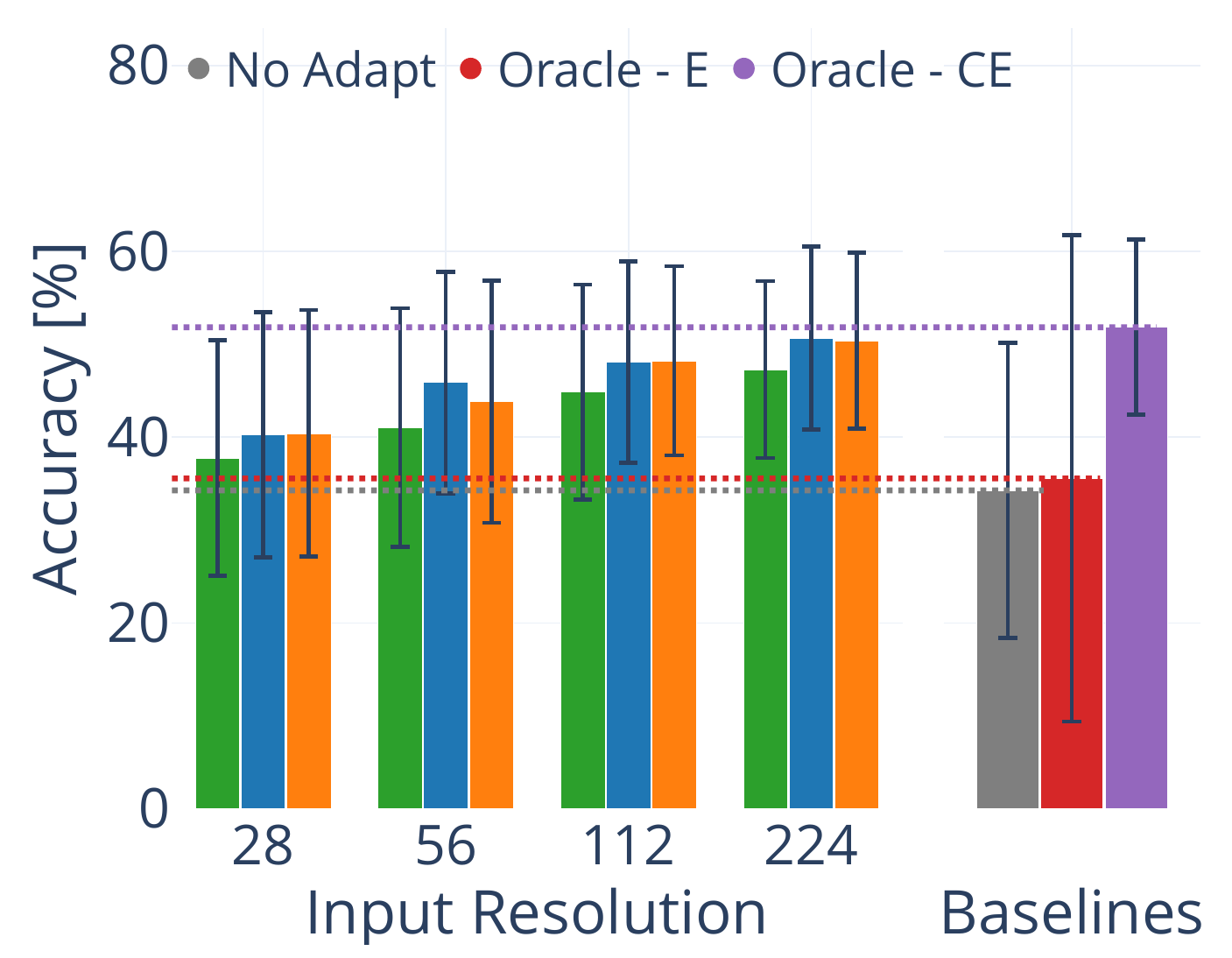}
  \caption{DeiT-Tiny.}
  \label{fig:bars_deit}
\end{subfigure}\hfill
\begin{subfigure}[t]{0.33\textwidth}
  \centering
  \includegraphics[width=\textwidth]{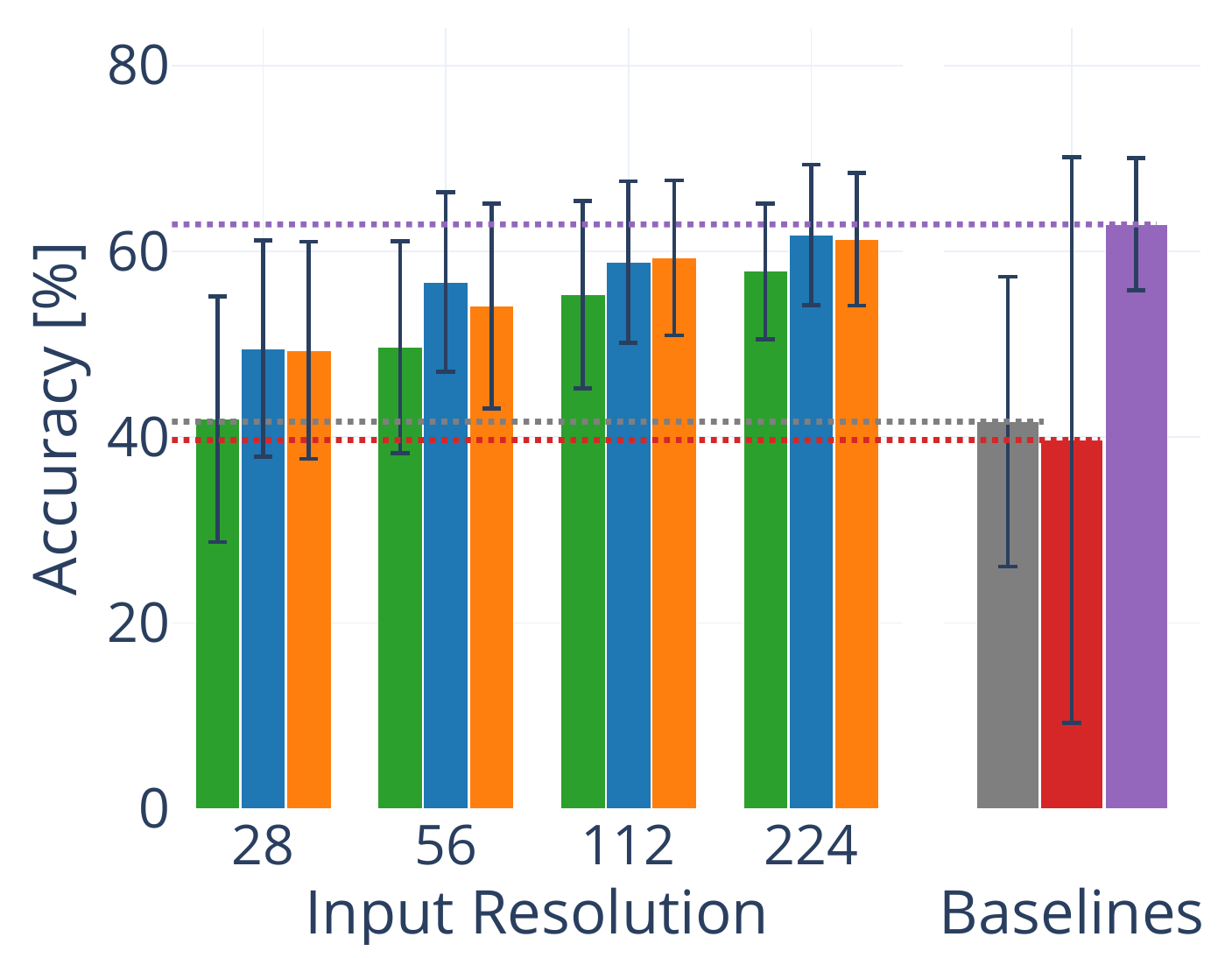}
  \caption{TinyViT-5M.}
  \label{fig:bars_tinyvit}
\end{subfigure}
\caption{Per-architecture accuracy across four input resolutions ($28$, $56$, $112$, $224$), reported for each frequency descriptor (RAPSD, DCT and DWT). \nfjf is compared against three resolution-independent references: \textit{No Adapt} and \textit{Oracle - E} baselines and the \textit{Oracle - CE} empirical upper bound. Bars report the mean over the $63$ test \cdag{}s of ImageNet-C; error bars indicate $\pm\sigma$. \textit{Oracle - E} accuracy markedly improves on \textit{NoAdapt} only for ConvNeXtV2-Femto and destabilizes on DeiT-Tiny and TinyViT-5M, confirming the fragility of unsupervised entropy-based adaptation under distribution shift. \nfjf accuracy improves on \textit{No Adapt} across all three architectures for resolutions $>28$ and it surpasses \textit{Oracle - E} at all resolutions on DeiT-Tiny and TinyViT-5M. For ConvNeXtV2-Femto, \nfjf mostly surpasses \textit{Oracle - E} at $112$ and $224$ input resolutions.}
\label{fig:bars_per_arch}
\end{figure*}

\circled{5} The host device adapts a replica of $f_\theta$ using the synthetic data by minimizing a pre-defined supervised loss over the compact parameter subspace of Sec.~\ref{sec:pwtta}, while all other model parameters remain frozen. 

\circled{6} The adapted parameters $\theta^*$ are transmitted back to the smart glasses, which replace the corresponding parameters in their local replica $f_\theta$, completing the adaptation cycle. Since only a parameter subset is exchanged, the update payload remains small enough for bandwidth-constrained links such as Bluetooth Low Energy. 

\section{Methodology}
\label{sec:method}

\subsection{Frequency Descriptors}
\label{sec:descriptors}

\begin{figure}[t]
\centering

\begin{subfigure}[t]{0.32\linewidth}
  \includegraphics[width=\textwidth]{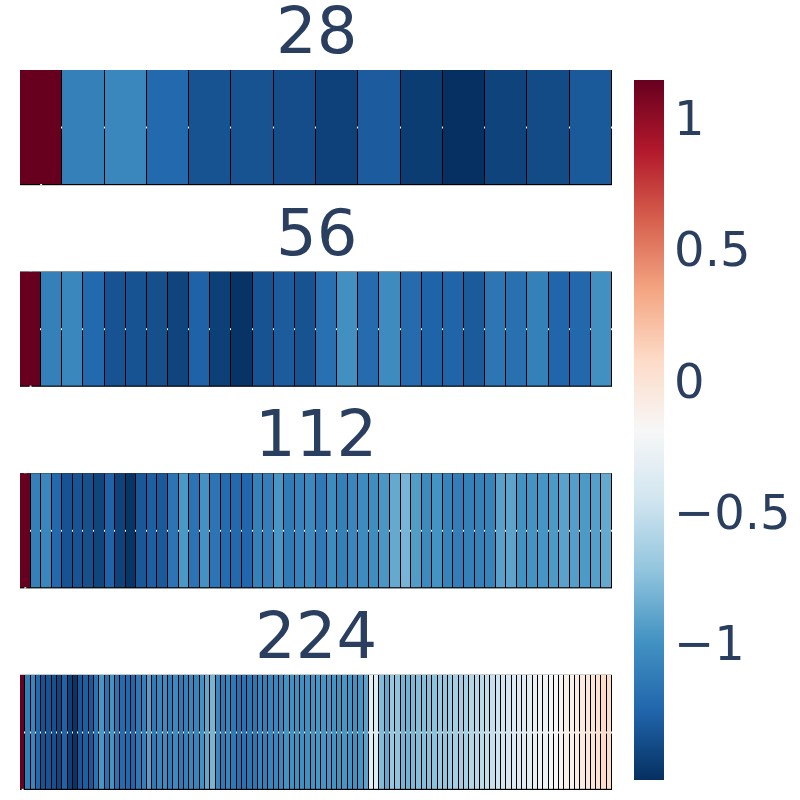}
  \caption{RAPSD}
  \label{fig:rapsd_vis}
\end{subfigure}\hfill
\begin{subfigure}[t]{0.32\linewidth}
  \includegraphics[width=\textwidth]{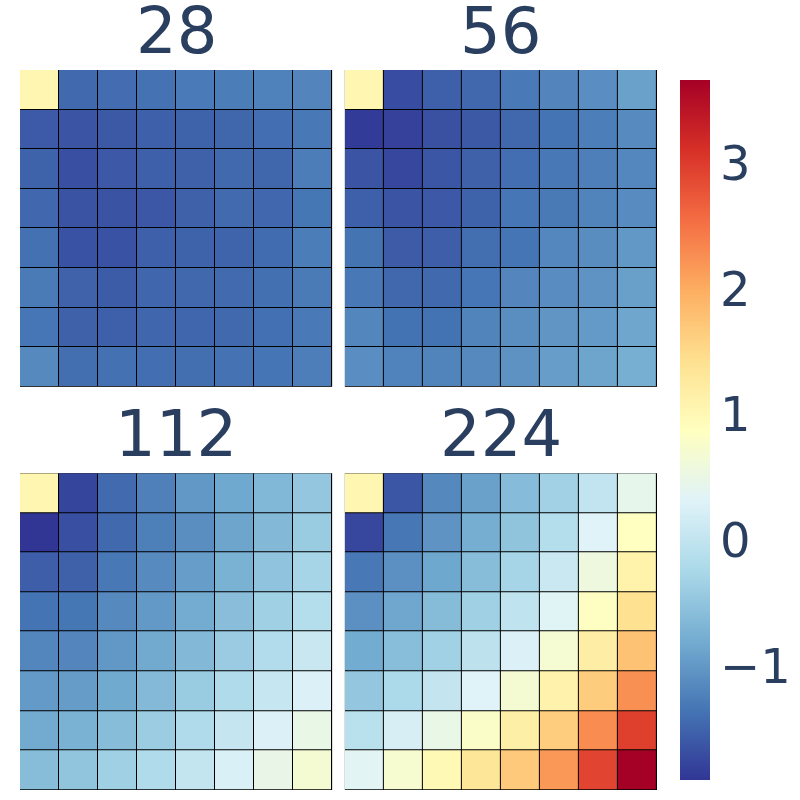}
  \caption{DCT}
  \label{fig:dct_vis}
\end{subfigure}\hfill
\begin{subfigure}[t]{0.32\linewidth}
  \includegraphics[width=\textwidth]{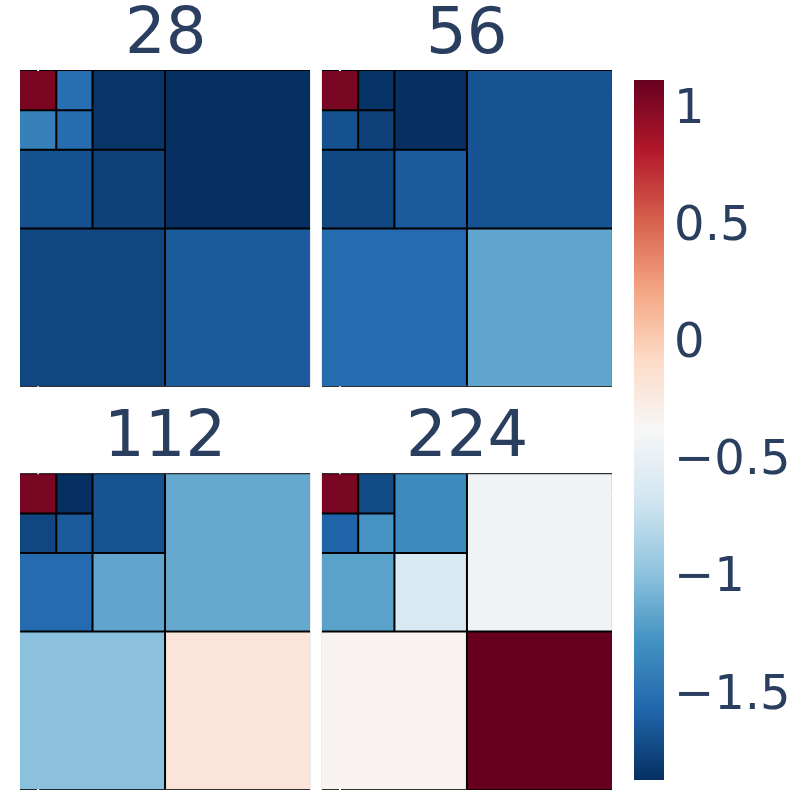}
  \caption{DWT}
  \label{fig:dwt_vis}
\end{subfigure}
\caption{
Descriptor visualizations under a sample \cdag (brightness $\rightarrow$ shot noise $\rightarrow$ contrast, each at severity 2; Section~\ref{sec:dags}), normalized with clean-data statistics. Brightness amplifies the DC component across all descriptors and resolutions; shot noise amplifies high frequencies, an effect that diminishes at lower resolutions since the reduced-scale JPEG decode acts as a low-pass filter. Contrast reduces the mid-frequency bands left free by the other two corruptions.}
\label{fig:descriptor_vis}
\end{figure}

\paragraph{Radially Averaged Power Spectral Density (\rapsd)} 

The Radially Averaged Power Spectral Density~\cite{van1996modelling} characterizes the global distribution of signal energy across spatial frequencies.  We compute \rapsd by applying a two-dimensional Fast Fourier Transform to the input image and averaging the power spectral density within concentric rings of increasing radius from the zero-frequency (DC) component, normalized by the number of pixels. This maps the two-dimensional spectrum to a compact one-dimensional vector of length $N/2$ (for an $N \times N$ input), where each index corresponds to a spatial frequency band. By explicitly discarding phase and directional magnitude information, this representation becomes rotation-invariant. 

\paragraph{Discrete Cosine Transform Statistics (\dct)}
The Discrete Cosine Transform~\cite{ahmed1974discrete} descriptor relies on the block-based frequency decomposition found in standard image compression pipelines (\ie JPEG, MPEG)~\cite{wallace1992jpeg}. We partition the input image into non-overlapping $8\times 8$ blocks and apply the two-dimensional DCT-II to each block via separable matrix multiplication using an orthonormal basis, and aggregate the coefficients across all blocks into the mean absolute value and standard deviation for each frequency index. This descriptor can be computed from the codec's pipeline. In contrast to \rapsd, this block-wise approach captures direction-specific texture signatures. 

\paragraph{Discrete Wavelet Transform Energy (\dwt)}
The Discrete Wavelet Transform (\dwt) captures frequency characteristics across multiple spatial scales~\cite{mallat1989wavelet}. We use the Haar wavelet for its computational efficiency on edge hardware~\cite{konecny2024computational}, with three-level recursive decomposition ($L=3$); at each level, the image is decomposed into the approximation (LL), horizontal (LH), vertical (HL), and diagonal (HH) detail sub-bands. We retain the three detail sub-bands (LH, HL, HH) as well as the approximation sub-band (LL), at each level. For each sub-band, we compute four statistics  (mean square energy, standard deviation, mean magnitude and peak value), yielding a fixed-size descriptor of $48$ dimensions ($3$ levels $\times$ $4$ sub-bands $\times$ $4$ stats), regardless of the input resolution. \dwt is part of the JPEG2000 standard~\cite{skodras2001jpeg2000}, thus these features could be extracted from existing encoder pipelines. 

All descriptors are computed on a single channel (red) of the incoming frame, to which the reported payload sizes refer; Sec.~\ref{sec:privacy} quantifies the personal information that these descriptors leak.

\subsection{Multi-Stage Corruption Classifier}
\label{sec:classifier}

Following the causal-graph corruption synthesis protocol of~\cite{drenkow2026causality}, we model each test-time visual distribution shift as a four-stage \cdag: $\mathcal{G} = (s_1, \ell_1, s_2, \ell_2, s_3, \ell_3, s_4, \ell_4)$, for each stage $s$ one corruption $c_s \in \mathcal{C}_s$ at severity $\ell_s \in \{0, 1, 2\}$ is sampled. The four stages follow the family sequence $\mathcal{K} = \{ \text{Weather}, \text{Blur}, \text{Noise}, \text{Digital} \}$ following~\cite{drenkow2026causality}, mimicking the chronological order in which real-world visual degradations arise~\cite{wang2021realesrgan}.

\paragraph{Architecture}
The classifier $g_\phi$ is a multi-head \mlp: a shared feature extractor of three fully connected layers followed by one linear head per stage. The head of stage $s$ outputs a distribution over the stage's corruption-severity pairs plus an inactive-stage class, yielding $3$, $5$, $9$, and $7$ classes for the four stages of \tabref{tab:dag_stages}. Descriptors are standardized with the statistics computed on the training split. Since the four stages differ in the number of predicted classes and in difficulty, their cross-entropy losses are combined through task-uncertainty weighting~\cite{kendall2018multi} rather than a plain sum, avoiding manual tuning of the four loss coefficients. One classifier is trained per descriptor type and input resolution using the same training configuration: AdamW, $100$ epochs, learning rate $5{\times}10^{-4}$, weight decay $10^{-4}$, batch size $200$, and a fixed random seed. The sensitivity of the adaptation gain to this seed is reported in Sec.~\ref{sec:class_multiseeds}.

\paragraph{Synthetic Mixture of Corruptions Pipeline}
The host maintains a pre-stored labeled set of clean adaptation images. This set is drawn from a range of validation indices strictly disjoint from the descriptor-training, descriptor-test and evaluation pools (Sec.~\ref{sec:dags}), preventing data leakage. Given a clean image $f_{clean}$ and the predicted per-stage corruption-severity distribution from the pre-trained classifier, the synthetic corrupted image is generated by chaining each predicted corruption operator in sequential order. 
Formally:
\begin{equation}
    f_{\text{synth}} = (\tilde{C}_{K} \circ \tilde{C}_{K-1} \circ \dots \circ \tilde{C}_{1})(f_{\text{clean}})
    \label{eq:sequential}
\end{equation}
where $\circ$ denotes function composition, and $\tilde{C}_{k}$ is the conditional corruption operator for the $k$-th stage.

\begin{table}[t]
    \centering
    \caption{\textbf{Four-stage ISP corruption \cdag.} Each \cdag is restricted to one corruption per node drawn from the disjoint set $\mathcal{C}_s$, a subset of ImageNet-C~\cite{hendrycks2019benchmarking} corruptions following~\cite{drenkow2026causality}. Each sample composes one corruption from each stage in the high-order degradation sequence following~\cite{wang2021realesrgan}.}
    \label{tab:dag_stages}
    \renewcommand{\arraystretch}{1.1}
    \setlength{\tabcolsep}{3pt}
    \footnotesize
    \resizebox{\columnwidth}{!}{%
        \begin{tabular}{c|l|l|c}
        \hline
        \textbf{Stage} & \textbf{Role} & \textbf{Pool $\mathcal{C}_s$} & \textbf{$|\mathcal{C}_s|$} \\
        \hline
        $S_1$ & Weather  & \{brightness\} & 1 \\
        $S_2$ & Blur     & \{defocus, gaussian\} blur & 2 \\
        $S_3$ & Noise    & \{gaussian, shot, speckle, impulse\} noise & 4 \\
        $S_4$ & Digital  & \{saturate, pixelate, contrast\} & 3 \\
    \hline
    \end{tabular}
    }
\end{table}

\subsection{Source-Assisted Adaptation Protocol}
\label{sec:pwtta}

Our proposed bandwidth-efficient adaptation strategy is based on the \tent framework~\cite{wang2020tent}, which adapts a model to data distribution shifts by updating only the affine parameters, $\gamma$ and $\beta$, of the normalization layers, while the other parameters remain fixed. \tent minimizes the prediction entropy on the test data on-the-fly, operating under the assumption that confident predictions correlate with test-time accuracy. The entropy loss $\mathcal{L}_{ent}$ is defined as:

\begin{equation}
    \label{eq:entropy}
    \mathcal{L}_{ent}(\hat{y}) = -\sum_{c} p(\hat{y}_{c}) \log p(\hat{y}_{c})
\end{equation}

\noindent where $p(\hat{y}_{c})$ is the model's predicted probability for class $c$. Entropy minimization can, however, be unstable, leading to model collapse~\cite{press2024entropy}. Since the host synthesizes the corrupted batches from pre-stored, labeled in-distribution data, ground-truth labels $y^{t}$ are available during adaptation, which enables the substitution of the unsupervised entropy loss with the cross-entropy loss:

\begin{equation}
    \label{eq:crossentropy}
    \mathcal{L}_{ce}(x^{t}, y^{t}) = -\sum_{c} y_{c}^{t} \log p(\hat{y}_{c})
\end{equation}

The cross-entropy loss yields a stronger supervisory signal than entropy minimization, which relies solely on prediction confidence and ignores class correctness~\cite{grandvalet2004semi}, and it does not admit the degenerate constant-prediction solution into which the entropy objective can collapse under severe distribution shift~\cite{press2024entropy}, an instability we observe for the entropy-based \textit{Oracle - E} baseline (Fig.~\ref{fig:bars_per_arch}). Leveraging the stability of the supervised objective, we extend the adapted parameter set to include the first block of each model~\cite{lee2023surgical}, which increases the adaptation accuracy gain: at $224$ resolution, removing the first block from this subspace costs $3.66{\pm}3.19$, $3.17{\pm}2.43$ and $1.15{\pm}2.22$\,pp on ConvNeXtV2-Femto, DeiT-Tiny and TinyViT-5M, and removing the normalization affine parameters costs $1.73{\pm}1.02$, $1.07{\pm}0.48$ and $2.96{\pm}1.41$\,pp; every entry is a mean ($\pm\sigma$) over the three descriptors and the $63$ test \cdag{}s. 

\begin{figure}[t]
\centering
\begin{subfigure}{\linewidth}
  \makebox[\linewidth][s]{\includegraphics[trim=14pt 0 0 0, clip, width=0.36\linewidth]{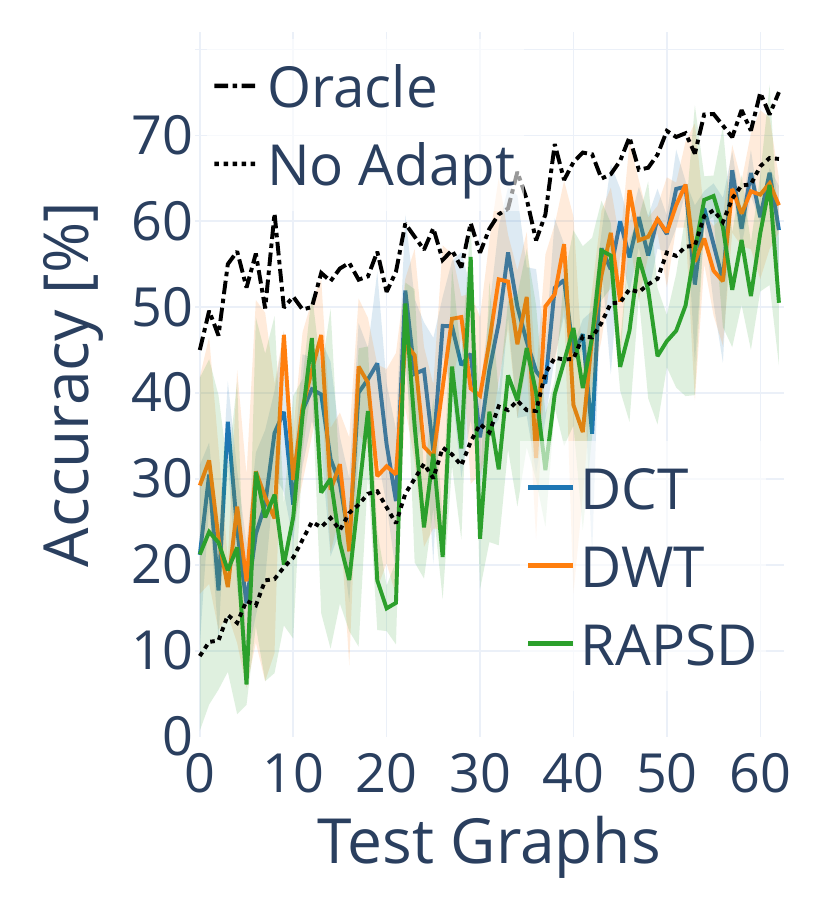}\hfill
  \includegraphics[width=0.297\linewidth]{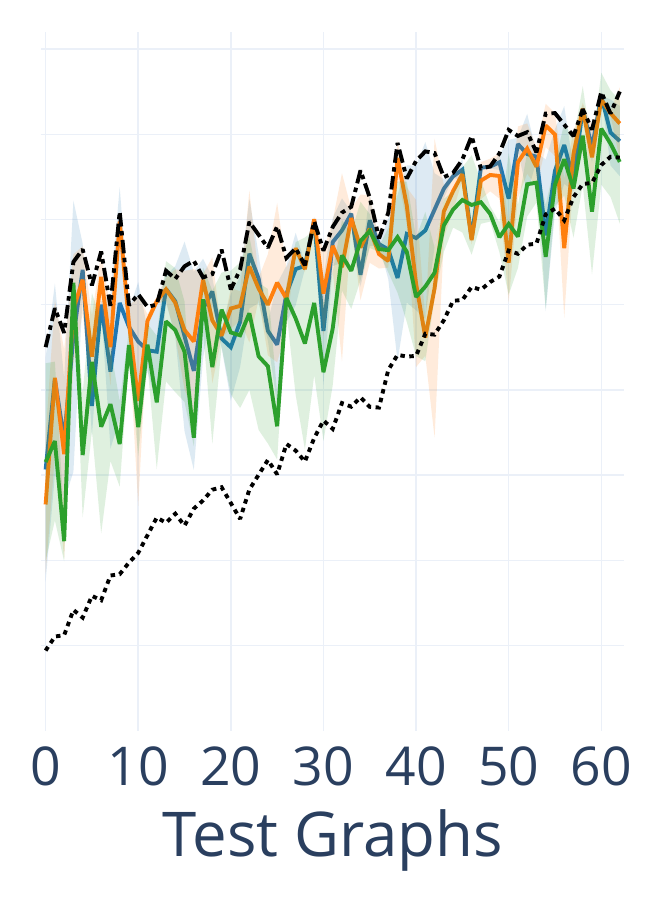}\hfill
  \includegraphics[width=0.297\linewidth]{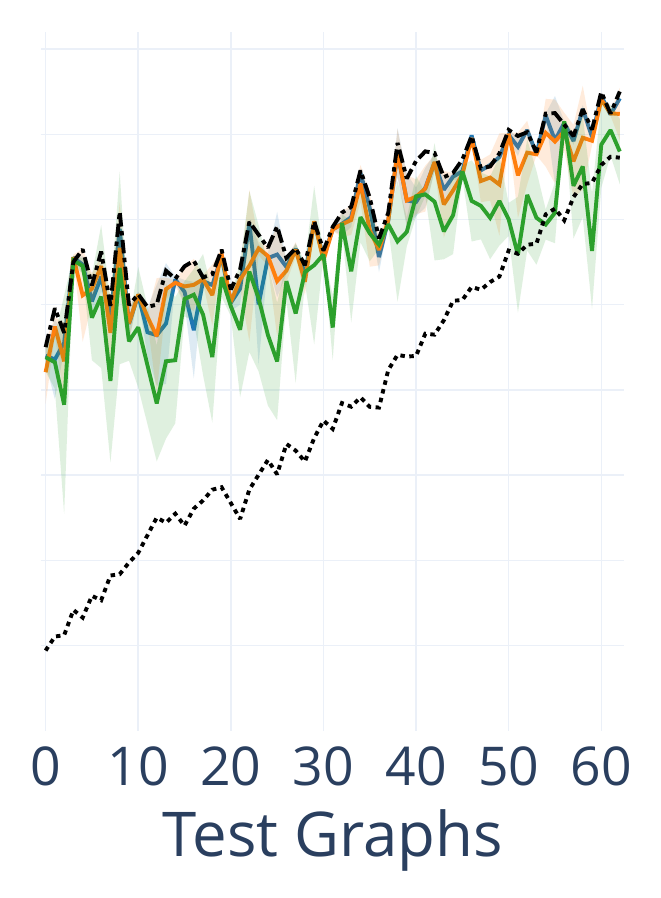}}
  \caption*{ConvNeXtV2-Femto.}
\end{subfigure}

\begin{subfigure}{\linewidth}
  \makebox[\linewidth][s]{\includegraphics[trim=14pt 0 0 0, clip, width=0.36\linewidth]{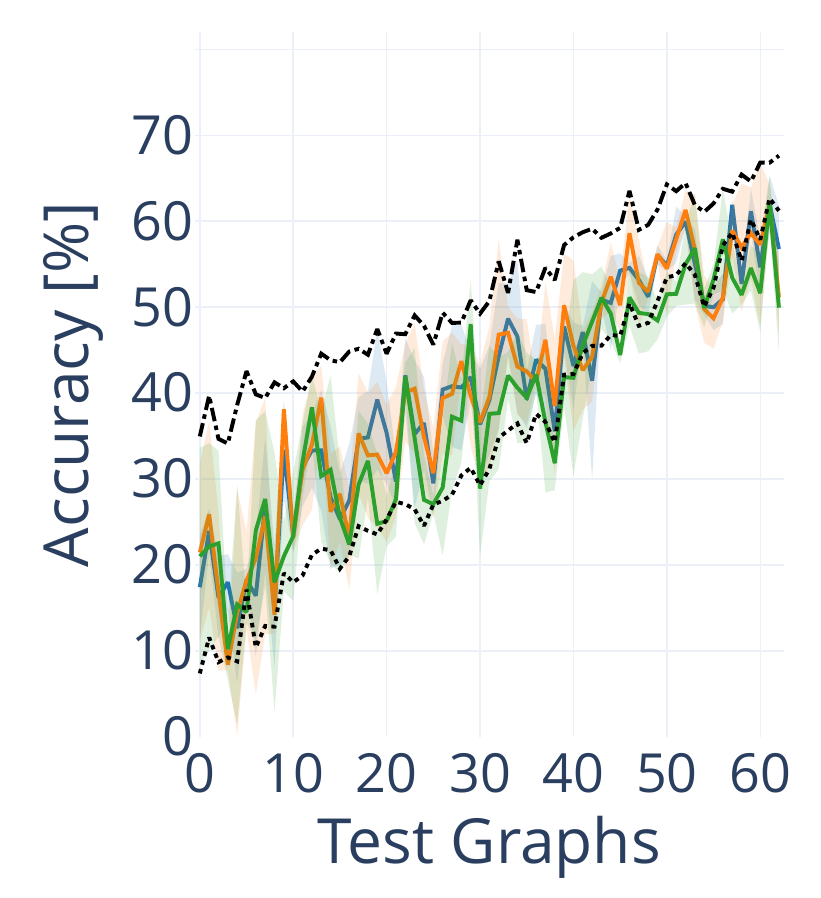}\hfill
  \includegraphics[width=0.297\linewidth]{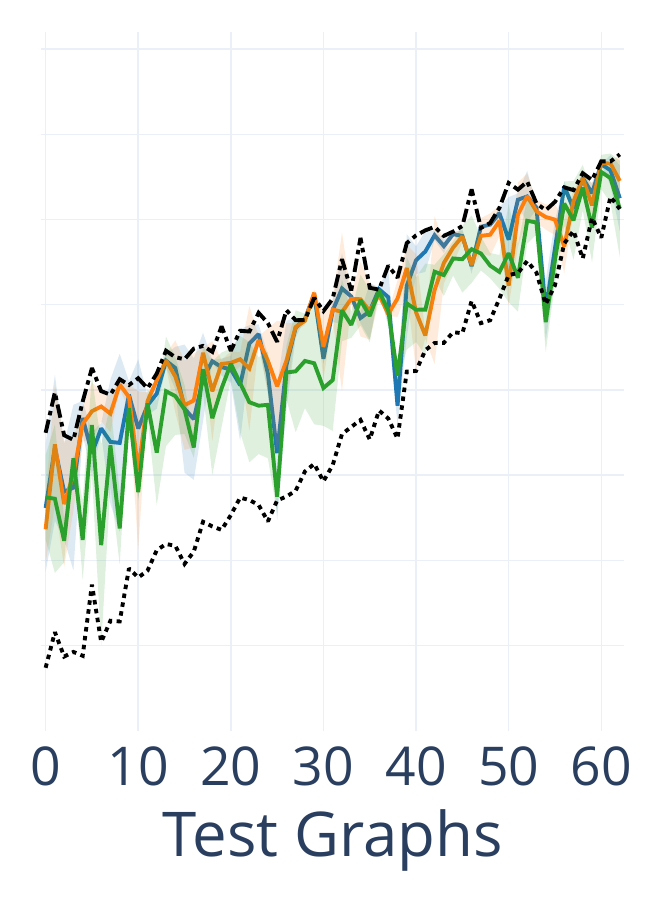}\hfill
  \includegraphics[width=0.297\linewidth]{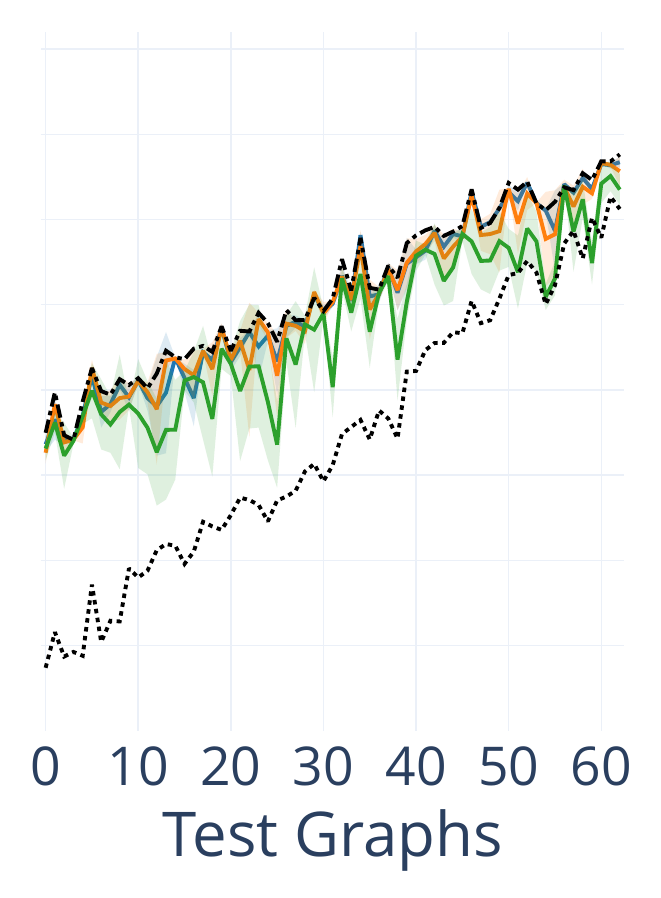}}
  \caption*{DeiT-Tiny.}
\end{subfigure}

\begin{subfigure}{\linewidth}
  \makebox[\linewidth][s]{\includegraphics[trim=14pt 0 0 0, clip, width=0.36\linewidth]{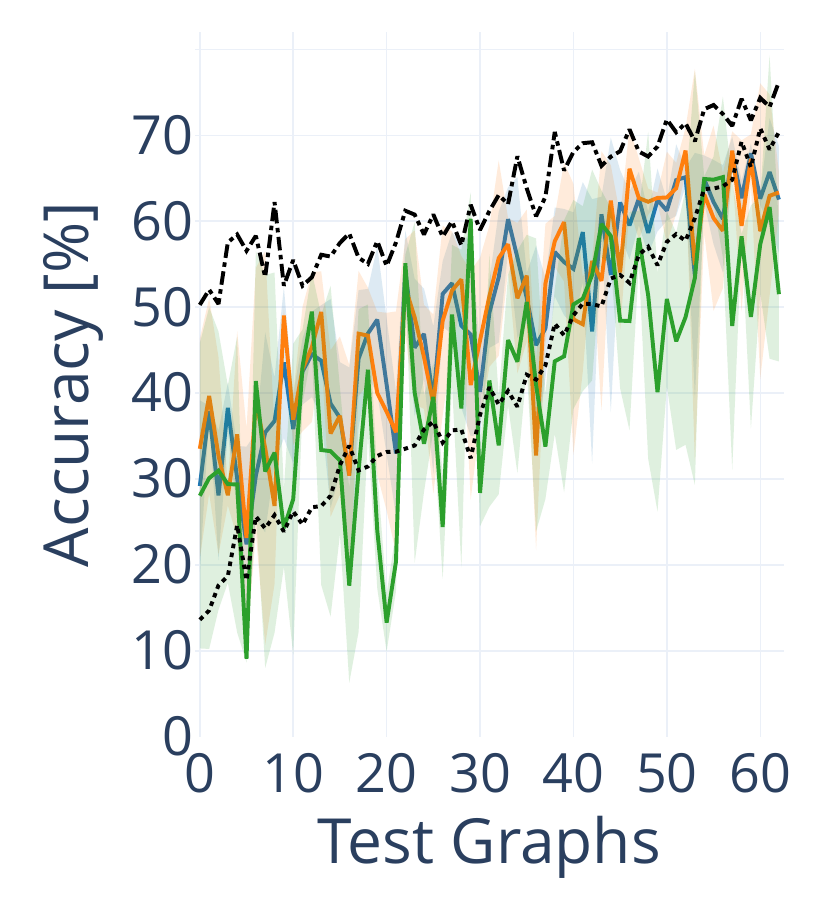}\hfill
  \includegraphics[width=0.297\linewidth]{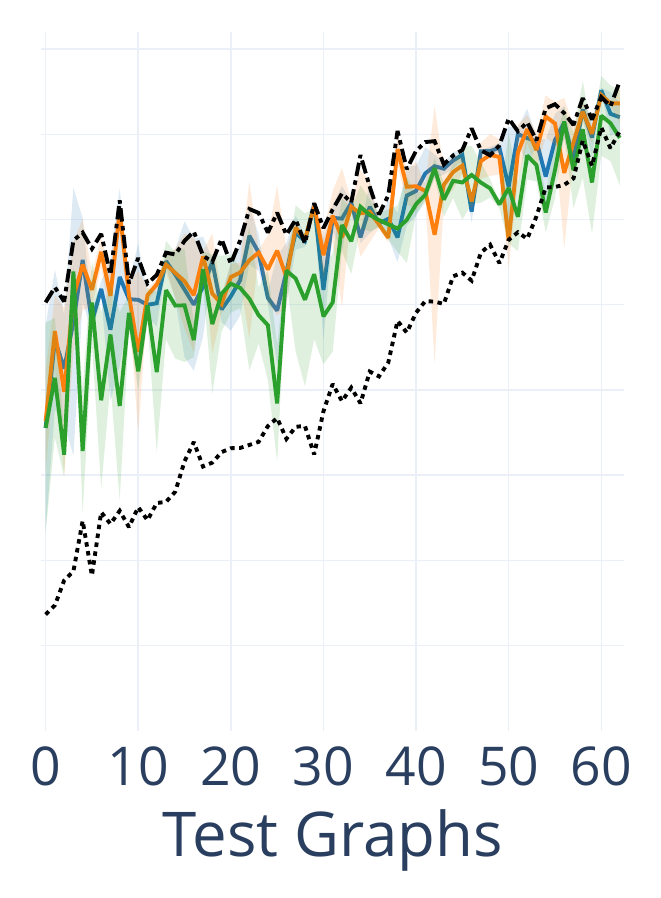}\hfill
  \includegraphics[width=0.297\linewidth]{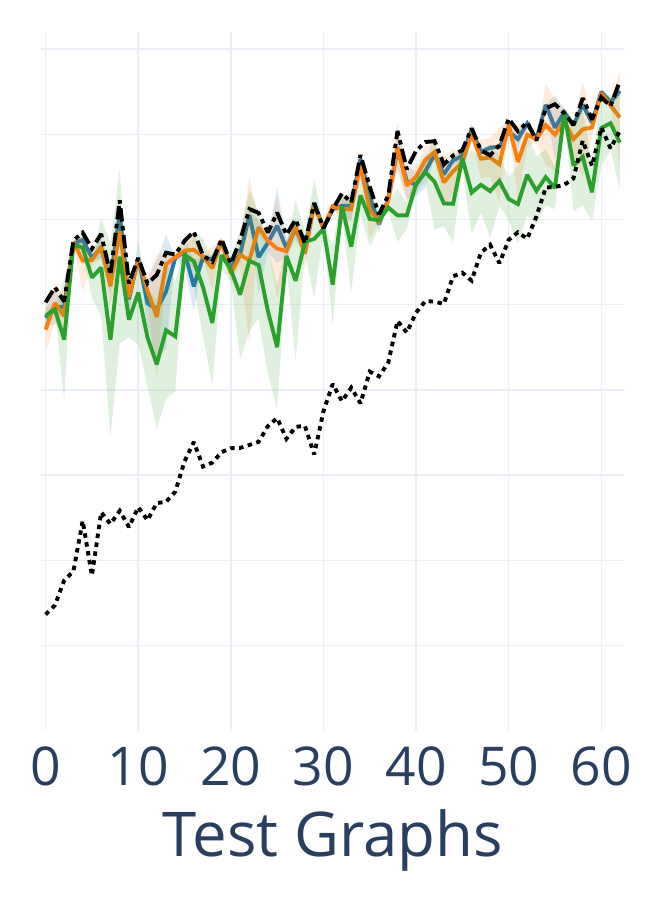}}
  \caption*{TinyViT-5M. }
\end{subfigure}
\caption{Adaptation accuracy for three architecture paradigms with increasing input resolutions across $G_{test}$. Left to right: $28$, $112$, $224$ input resolutions. The frequency descriptors DCT, DWT, and RAPSD (Section~\ref{sec:descriptors}) are compared against the \textit{No Adapt} and the \textit{Oracle} baselines. The test graphs are sorted along the x-axis in ascending order of their mean \textit{NoAdapt} accuracy across all architectures. 
As the input resolution increases the adaptation gap of \nfjf against supervised cross-entropy Oracle adaptation narrows, with fewer $G_{test}$ shifts falling below \textit{No Adapt}.}
\label{fig:per_graph_main_grid}
\end{figure}

\section{Experimental Setup}
\label{sec:experiments}

\subsection{Datasets and DAGs}
\label{sec:dags}

The multi-stage corruption classifier, $g_\phi$, is trained on a set of $945$ corruption mixtures, denoted as $G_{train}$ generated via the multi-stage \cdag model (Sec.~\ref{sec:classifier}).
To train the classifier (\circled{3}), these \cdag{}s are applied to the first image of each class within the ImageNet validation dataset~\cite{deng2009imagenet}.
Subsequently, the frequency descriptors are extracted, yielding 945,000 distinct frequency signatures per classifier. This process is repeated for every input resolution and descriptor type.

The test \cdag{}s (denoted as $G_{test}$) are a subset of the training mixtures ($G_{test} \subset G_{train}$); moreover to generate the evaluation data for the classifier, $G_{test}$ is applied to the second image of each class. This ensures a strict train-test descriptor separation while remaining within the trained corruption-mixture distribution.

\subsection{Adaptation Protocol}
\label{sec:adapt_protocol}
Ultimately, $G_{test}$ evaluates both the corruption classifier and the end-to-end adaptation pipeline performance. For the end-to-end evaluation, we partition the ImageNet validation dataset~\cite{deng2009imagenet} into datasets drawn from two disjoint image sets:
\begin{itemize}
    \item \textbf{$D_{eval}^{true}$:} Formed by applying the ground-truth corruption graphs from $G_{test}$ to the 3rd to 12th image of each class. This yields 10,000 images per graph, totaling 630,000 images.
    \item \textbf{$D_{synth}^{pred}$:} Constructed by applying the classifier's predicted corruptions to the 13th through 50th images of each class. This results in 38,000 images per graph, yielding a total of 2,394,000 images. The oracle baselines instead adapt on $D_{synth}^{true}$ produced by the same synthesis pipeline using the ground-truth test \cdag instead of the predicted one; these two datasets differ only in the applied corruption graph. 
\end{itemize}

This partitioning strictly separates the evaluation data (\mbox{$D_{eval}^{true}$}) from the \textit{synthetic} adaptation data (\mbox{$D_{synth}^{pred}$} and \mbox{$D_{synth}^{true}$}). The end-to-end adaptation is executed independently for each test \cdag, re-initializing each model to its pre-trained parameters. For every test \cdag and descriptor configuration, the full adaptation cycle is executed three times with independently sampled descriptors and synthetic batches. For each test \cdag the reported value is the mean over these three repetitions; the error bars in all figures indicate the standard deviation of these means across the $63$ test \cdag{}s. 

\subsection{Evaluation Baselines}
\label{sec:baselines}
\textbf{No Adapt (Source-Only).}  The pre-trained model $f_\theta$ is evaluated on the corrupted evaluation data $D_{eval}^{true}$ \emph{without} any parameter update. 
This baseline quantifies the accuracy degradation induced by the distribution shift and provides the common reference for the accuracy gain of every adaptation method considered.

\textbf{Oracle - E (Entropy Minimization).}  Following Wang \etal~\cite{wang2020tent}, we minimize the entropy of the model's softmax predictions on $D_{synth}^{true}$, matching the adaptation budget of \textit{Oracle - CE}. In line with the adaptation used in our method, the affine parameters of the normalization layers ($\gamma$ and $\beta$), as well as the first block~\cite{lee2023surgical}, are adapted, while all other parameters remain frozen. This baseline serves as the reference for on-device unsupervised adaptation and requires access to all the corrupted data, unrealistic for edge deep learning inference frameworks, which lack gradient computation~\cite{lin2023tiny}.

\textbf{Oracle - CE (Supervised Cross Entropy).} 
The Oracle adapts the model on $D_{synth}^{true}$ using ground-truth labels via cross-entropy loss (Eq.~\ref{eq:crossentropy}), updating affine parameters of the normalization layers, as well as the model's first block~\cite{lee2023surgical}. 
Since it has access to both the ground-truth \cdag and the corresponding labels, Oracle - CE serves as the empirical upper bound on the adaptation gain within our adaptation framework, and its gap to \nfjf isolates the impact of predicting the corruption graph at test time. For the object detection task, the supervised oracle's adaptation loss (\textit{Oracle - CE + Reg.}) contains, in addition to the cross-entropy term, the bounding-box regression term, which penalizes the localization error between the predicted and the ground-truth bounding boxes. 


\subsection{Experiments}

\textbf{Architectures.}
To assess the scalability of the framework across different classification models, we evaluate three architectures that span convolution, self-attention and hybrid designs: \emph{(i)} ConvNeXtV2-Femto~\cite{woo2023convnext}, a fully convolutional backbone with Layer Normalization; 
\emph{(ii)} DeiT-Tiny~\cite{touvron2021deit}, a Vision Transformer backbone; 
\emph{(iii)} TinyViT-5M~\cite{wu2022tinyvit}, a hybrid convolution-transformer backbone using Batch Normalization in its convolution layers and Layer Normalization in its self-attention layers. All models contain approximately five million parameters, so that accuracy differences across the three paradigms reflect architectural design rather than scale. Following Lee \etal~\cite{lee2023surgical}, we use Adam~\cite{kingma2014adam}, learning rate $10^{-4}$, weight decay $10^{-4}$ and batch size 64. 


\begin{table*}[t]
\caption{Framework generality across person classification and object detection. Entries report the mean (box mAP on MS-COCO; accuracy [\%] on Visual Wake Words) per-\cdag paired gain ($\pm\sigma$) over the $63$ test \cdag{}s. The \textit{No Adapt} row  reports the absolute performance under shift, the common reference of every adaptation gain. The best per-column adaptation gain is reported in bold. The trends observed in Fig.~\ref{fig:bars_per_arch} persist: every adaptation gain is positive and approaches the supervised \textit{Oracle} as the resolution increases, whereas \textit{Oracle - E} (VWW) collapses below \textit{No Adapt}.}
\label{tab:taskgen}
\centering
\footnotesize
\begin{tabular}{l|cccc|cccc|ccc}
\toprule
 & \multicolumn{4}{c|}{\textbf{SSDLite320-MNv3-L}} & \multicolumn{4}{c|}{\textbf{YOLOX-Tiny}} & \multicolumn{3}{c}{\textbf{MobileNetV1-0.25, VWW}} \\
\midrule
\textit{No Adapt} & \multicolumn{4}{c|}{$7.4{\pm}5.1$} & \multicolumn{4}{c|}{$8.8{\pm}7.1$} & \multicolumn{3}{c}{$65.3{\pm}4.2$} \\
\textit{Oracle} & \multicolumn{4}{c|}{$+4.7{\pm}1.4$ (CE + Reg.)} & \multicolumn{4}{c|}{$+8.3{\pm}2.2$ (CE + Reg.)} & \multicolumn{3}{c}{$+9.1{\pm}3.2$ (CE); \; $-2.6{\pm}4.2$ (E)} \\
\midrule
Input Resolution & 40 & 80 & 160 & 320 & 52 & 104 & 208 & 416 & 24 & 48 & 96 \\
\midrule
\rapsd & $+1.6{\pm}1.9$ & $+2.0{\pm}1.6$ & $+2.6{\pm}1.8$ & $+3.7{\pm}1.4$ & $+3.3{\pm}3.1$ & $+4.8{\pm}2.5$ & $+5.4{\pm}2.8$ & $+6.8{\pm}2.2$ & $+4.3{\pm}3.4$ & $+5.8{\pm}3.5$ & $+5.8{\pm}3.7$ \\
\dct   & $\mathbf{+2.1{\pm}1.5}$ & $+2.8{\pm}1.6$ & $+3.4{\pm}1.7$ & $+4.4{\pm}1.5$ & $+4.5{\pm}2.1$ & $\mathbf{+6.1{\pm}2.9}$ & $+6.9{\pm}2.5$ & $+8.0{\pm}2.4$ & $+5.8{\pm}3.6$ & $+6.5{\pm}3.2$ & $\mathbf{+7.5{\pm}3.2}$ \\
\dwt   & $+1.6{\pm}1.8$ & $\mathbf{+2.9{\pm}1.8}$ & $\mathbf{+3.9{\pm}1.8}$ & $\mathbf{+4.4{\pm}1.5}$ & $\mathbf{+4.6{\pm}2.8}$ & $+5.8{\pm}2.4$ & $\mathbf{+7.0{\pm}2.6}$ & $\mathbf{+8.1{\pm}2.3}$ & $\mathbf{+6.4{\pm}3.2}$ & $\mathbf{+7.1{\pm}3.4}$ & $+7.5{\pm}3.8$ \\
\bottomrule
\end{tabular}
\end{table*}

\textbf{Resolution Scaling.}
Since the cost of computing the descriptors depends on the input resolution, we evaluate for each test graph the impact of different input resolutions on the adaptation accuracy gain. Each corrupted frame is re-encoded in memory as a JPEG at the maximum quality factor ($Q{=}100$) and decoded directly at $1/8$, $1/4$, or $1/2$ in the \dct domain with the TJpgDec library~\cite{krumpl2026deploying}, mirroring the reduced-scale decode of embedded JPEG decoders; the maximum-quality setting minimizes compression artifacts. This spatial reduction acts as a low-pass filter (Fig.~\ref{fig:descriptor_vis}): the shot noise signature progressively diminishes as the resolution decreases, whereas contrast and brightness, which perturb the mid-to-low bands and the zero-frequency (DC) component, remain visible at every resolution scale. 

\subsection{Person Classification and Object Detection}

\textbf{Person Classification.}
We port the framework without modifications to Visual Wake Words~\cite{chowdhery2019visual} (VWW), a binary person/no-person classification task, with a MobileNetV1 ($\alpha = 0.25$, $96 \times 96$ input)~\cite{howard2017mobilenets}. The fractional split of Sec.~\ref{sec:dags} applied per class yields disjoint descriptor-classifier training (810) and test (810) datasets, an evaluation dataset of $8{,}101$ images and an adaptation dataset of $30{,}783$ images. The same $63$ test \cdag{}s are used, the descriptor resolutions follow the $96$ input-halving chain $\{24, 48, 96\}$, and the baselines of Sec.~\ref{sec:baselines} are reported at resolution $96$. This task serves the privacy leakage analysis of Sec.~\ref{sec:privacy}.   

\textbf{Object Detection.}
We further port the framework to object detection on MS-COCO~\cite{lin2014microsoft} with two models: SSDLite-MobileNetV3-Large~\cite{howard2019searching} ($3.4$M parameters, $320^2$ input) and YOLOX-Tiny~\cite{ge2021yolox} ($5.1$M parameters, $416^2$ input). We draw all the partition datasets from  \texttt{val2017}, since the released weights are trained on \texttt{train2017}. The $5{,}000$ images are partitioned using the fractional split of Sec.~\ref{sec:dags}. Since the corruptions of Eq.~\eqref{eq:sequential} are pixel-level operations that leave the image geometry unchanged, the adaptation dataset ground-truth boxes remain valid labels for the synthesized batches. The supervised Oracle is the \textit{Oracle - CE + Reg.} variant of Sec.~\ref{sec:baselines}. The adapted subspace mirrors the classification one; we use Adam with learning rate $10^{-4}$, batch size 16 and a single epoch. We report the COCO box mAP[.50:.95]~\cite{lin2014microsoft} computed on the evaluation dataset. 

\begin{figure*}[t]
\centering
\begin{subfigure}[t]{0.3\textwidth}
  \centering
  \includegraphics[width=\textwidth]{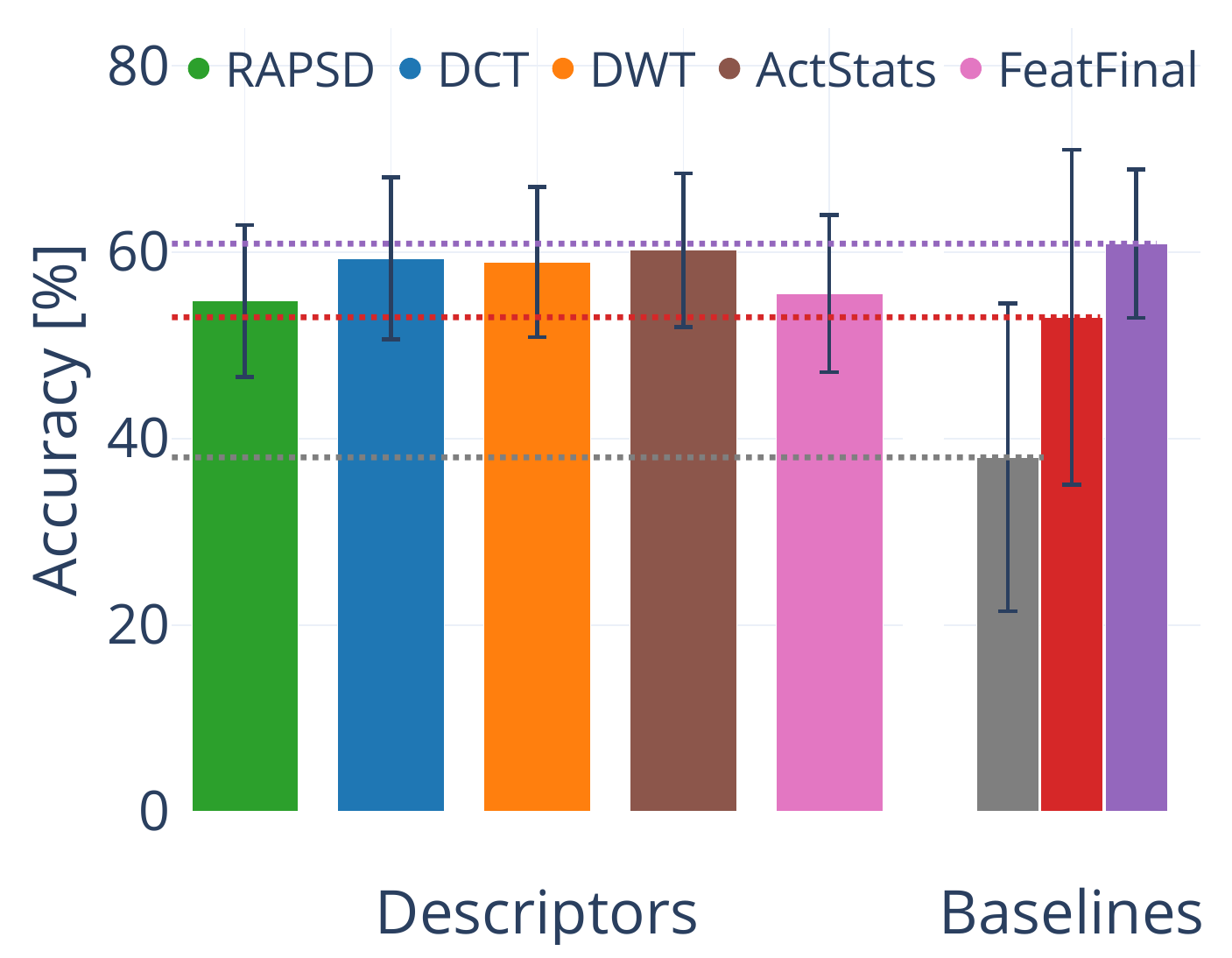}
  \caption{ConvNeXtV2-Femto.}
  \label{fig:ablation_convnext}
\end{subfigure}\hfill
\begin{subfigure}[t]{0.3\textwidth}
  \centering
  \includegraphics[width=\textwidth]{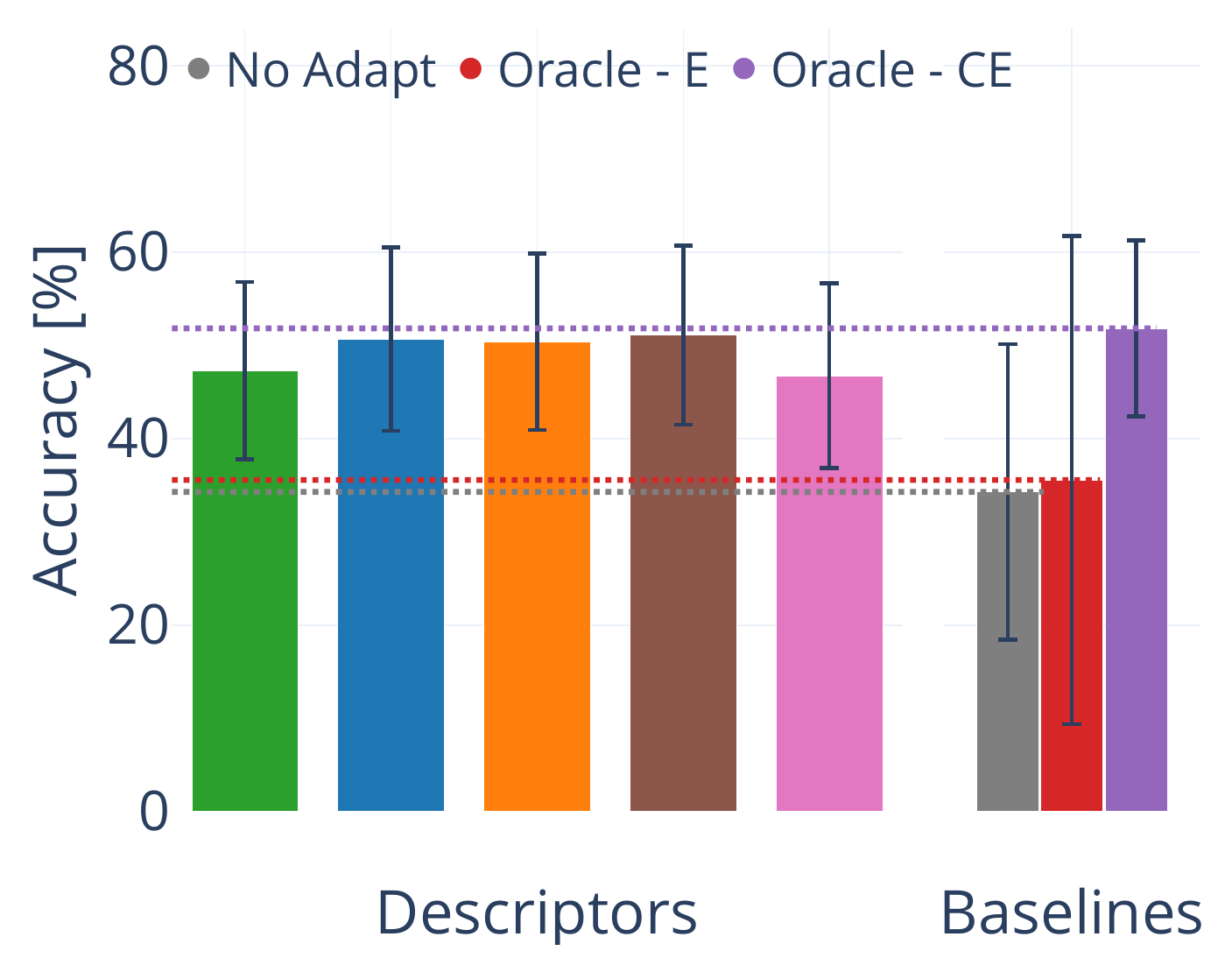}
  \caption{DeiT-Tiny.}
  \label{fig:ablation_deit}
\end{subfigure}\hfill
\begin{subfigure}[t]{0.3\textwidth}
  \centering
  \includegraphics[width=\textwidth]{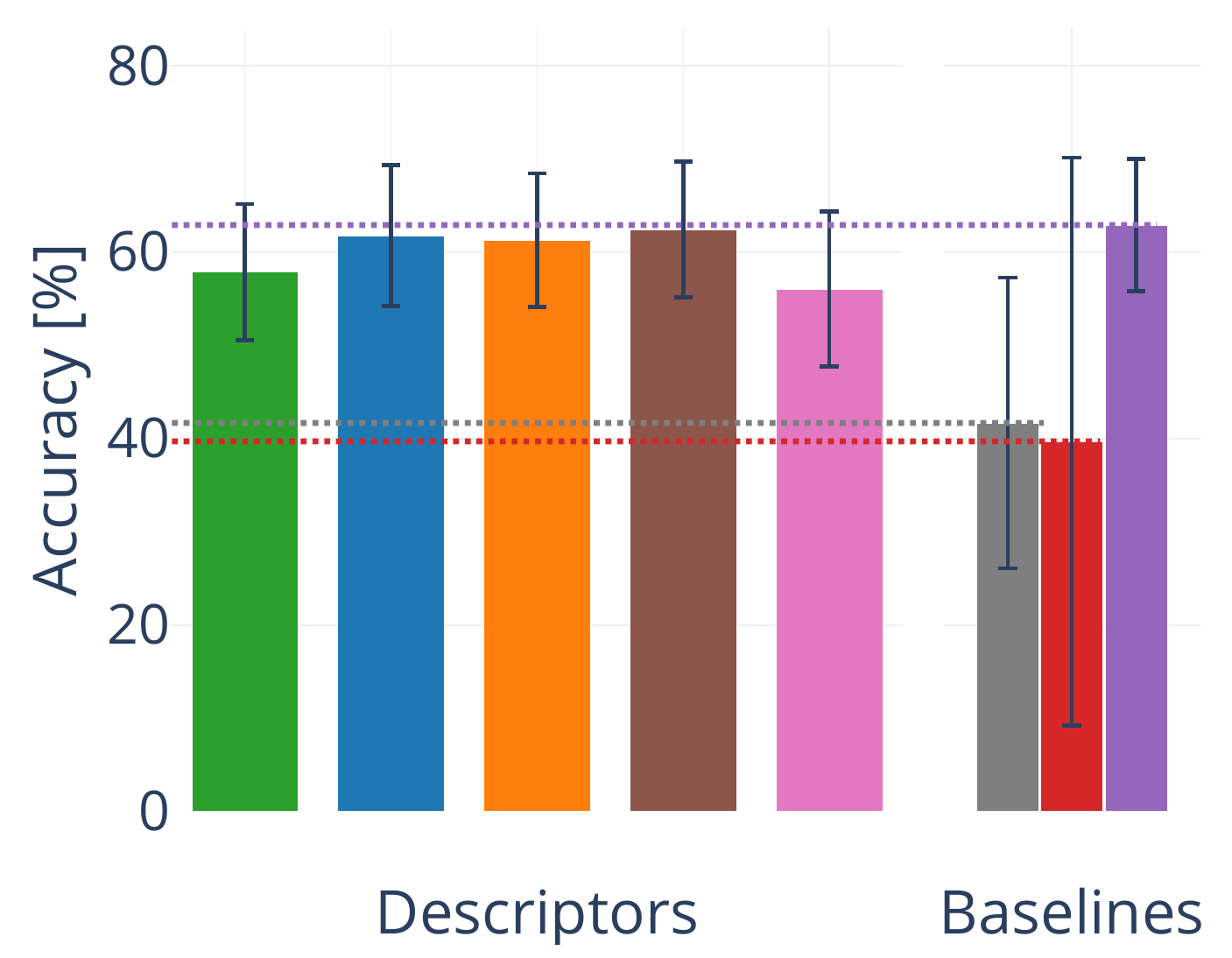}
  \caption{TinyViT-5M.}
  \label{fig:ablation_tinyvit}
\end{subfigure}
\caption{Descriptors ablation at $224$ input resolution: the \nfjf adaptation loop re-run with the frequency descriptor replaced by the model-based alternatives. \textit{ActStats} denotes the first layer and block activation statistics and \textit{FeatFinal} the pooled pre-logits embedding. Same layout as Fig.~\ref{fig:bars_per_arch}. \textit{ActStats} attains the highest adaptation accuracy, improving on \dct by $0.67$\,pp on average and approaching \textit{Oracle - CE} within $0.71$\,pp. This adaptation gain costs on average $2.3\times$ the \dct payload (a fixed $708$\,B versus architecture-dependent $768$--$3072$\,B). Computing \textit{ActStats} requires executing the deployed model's first layer and block, $\approx100\times$ the number of \dct FLOPs ($\sim150$-$180$ versus $1.6$ MFLOPs), whereas the frequency descriptors need no model execution.}
\label{fig:desc_ablation}
\end{figure*}

\section{Results} 

\subsection{Adaptation Across Architectures and Resolutions}

On ImageNet-C (Fig.~\ref{fig:bars_per_arch}, Fig.~\ref{fig:per_graph_main_grid}), we observe that \nfjf recovers most of the supervised oracle adaptation gain. At the native 224 descriptor resolution, the strongest descriptor (DCT) recovers $93.1\%$ of the gain on ConvNeXtV2-Femto ($59.34\%$ versus \textit{Oracle - CE} $60.91\%$, from a \textit{No Adapt} accuracy of $37.99\%$), $93.3\%$ on DeiT-Tiny, and $94.7\%$ on TinyViT-5M. In contrast, \textit{Oracle - E}, despite consuming the full corrupted stream ($D_{synth}^{true}$), reports a notable adaptation gain only on the convolution architecture and shows adaptation collapse in the self-attention and hybrid architectures, the known degeneracy of the entropy objective into a trivial constant-prediction solution~\cite{press2024entropy, niu2023towards}. For every architecture the adaptation gains grow monotonically with the descriptor resolution (Fig.~\ref{fig:bars_per_arch}).

\subsection{Classifier Seed Sensitivity}
\label{sec:class_multiseeds}

We train two additional corruption classifiers per descriptor and input resolution with different random seeds and repeat the ImageNet-C adaptation loop with each classifier on one transmitted descriptor per test \cdag. We observe that the three classifiers predict the same \cdag in 4\% of the test cases at resolution 28 and in 59\% at 224, whereas the standard deviation of the post-adaptation accuracy across the three classifiers ($4.4$, $3.4$, $1.6$ and $0.7$\,pp at $28$, $56$, $112$ and $224$) falls below the standard deviation across the three transmitted descriptors of the same \cdag ($7.3$, $5.2$, $3.8$ and $2.5$\,pp; Sec.~\ref{sec:descriptors}). The adaptation gain increases monotonically with the resolution for every classifier; thus the trends observed in Fig.~\ref{fig:bars_per_arch} do not depend on the classifier seed, and the transmitted frame, rather than the classifier, is the major source of adaptation variability.  

\subsection{Task Generality}

\textbf{Person Classification.} On VWW (\tabref{tab:taskgen}), we observe that \nfjf improves the corrupted data accuracy from $65.3\%$ to $72.80\%$ (\dwt at $96$ resolution), closing $82\%$ of the $9.1\%$ adaptation gain of \textit{Oracle - CE}, whereas \textit{Oracle - E} falls below \textit{No Adapt}. We note that the binary task compresses the adaptation gap range: the accuracy under shift is bounded by \textit{No Adapt} at $65.3$\% on average, thus the margin recoverable through adaptation is correspondingly small. We believe the fraction of the \nfjf versus oracle adaptation gap is therefore a more informative statistic than the adaptation gain. 

\textbf{Object Detection.} On MS-COCO (\tabref{tab:taskgen}), we observe that \nfjf improves box mAP from $7.4$ to $11.8$ on SSDLite320 (\dwt at 320 input resolution), against an oracle adaptation accuracy of $12.1$ ($93.6$\% of the achievable gain), and from $8.8$ to $16.9$ on YOLOX-Tiny against $17.1$ ($97$\%). The adaptation gap closure is highest on detection compared to classification, consistent with the denser per-image supervision detection objective (\ie per-anchor classification and regression ground-truth boxes).  

\subsection{Descriptor Ablation}
\label{sec:desc_ablation}

To isolate the contribution of the frequency descriptor, we re-run the adaptation loop at the native $224$ resolution with the descriptor replaced by model-based alternatives (Fig.~\ref{fig:desc_ablation}): \textit{ActStats}, the first layer and block activation statistics, and \textit{FeatFinal}, the pre-logits embedding. We observe that \textit{ActStats} attains the highest adaptation accuracy on every architecture, improving on \dct by $0.67$\,pp on average and approaching \textit{Oracle - CE} to within $0.63$\,pp on average, whereas \textit{FeatFinal} falls below \dct and \dwt on every architecture, $5.1$\,pp below \textit{ActStats} on average: the corruption \cdag is better identified from the low-level spectral content preserved by the first layers rather than from the pre-logits embedding. We believe the frequency descriptors remain the better choice under deployment constraints, since the $0.67$\,pp adaptation increase of \textit{ActStats} costs on average $2.3\times$ the \dct payload and $\approx100\times$ its FLOPs.

\subsection{Privacy Leakage Analysis}
\label{sec:privacy}
We consider an eavesdropper on the wireless uplink that observes only the transmitted frequency descriptors. We measure the leakage with a classifier that predicts a sensitive attribute, the person-presence label of VWW, from the single descriptor~\cite{malekzadeh2019mobile}. We split the descriptor-classifier test images into training and test (80:20), apply the $63$ test \cdag{}s, train a logistic regression, a gradient-boosted tree and an \mlp classifier. We repeat the split five times and report the mean ($\pm\sigma$) of the most accurate of the three classifiers. We use the balanced accuracy, \ie the mean of the person and no-person accuracies, since VWW is not a balanced dataset. We observe that the most accurate eavesdropper obtains $58.5{\pm}2.4\%$ (\dct at resolution $48$), thus the descriptors carry a weak person-presence signal rather than disclosing the attribute. We note that the leakage is lowest at the lowest descriptor resolution for every descriptor and that it saturates as the resolution grows: averaged over the three descriptors, raising the resolution from $24$ to $48$ increases the adaptation gain by $1.0$\,pp and the leakage by $1.3$\,pp, whereas raising it from $48$ to $96$ increases the gain by $0.4$\,pp and the leakage by only $0.3$\,pp. We believe the descriptor resolution therefore trades adaptation gain for leakage mainly at the lowest resolutions range. 

\subsection{Adaptation Latency on Smart Glasses}
We measure the end-to-end latency of the adaptation loop on a Seeed Grove Vision AI V2 board (Himax WiseEye2, Cortex-M55), the processor class of the Brilliant Labs Halo smart glasses~\cite{brilliant2026halo}, running the VWW MobileNetV1 architecture as an int8 model~\cite{david2020tensorflow}. Over $10$ adaptation runs (\dct resolution 96, laptop CPU host), one adaptation requires $85{\pm}9$\,s: the synthesis of the image adaptation set ($41{\pm}3$\,s) and the cross-entropy forward-backward pass ($44{\pm}7$\,s) account for 98\% of the time. The descriptor computation, the update of the adapted subspace in SRAM and the first adapted inference require $38$\,ms on the device. The transfer of the $708$\,B descriptor and of the $22$\,kB update requires $0.14$\,s at the 1.32 Mbit/s measured for BLE 5~\cite{wang2020ble}. On a Samsung Galaxy S23 (Snapdragon 8 Gen 2) using four CPU cores, the host passes require $84{\pm}9$\,s and $218{\pm}7$\,s over $5$ runs. The energy consumption of the host stages and their mobile GPU implementation, for which to the best of our knowledge no off-the-shelf framework exists, are left for future work.

\section{Discussion} 
This paper analyzes whether deployment-time model adaptation for smart glasses can be driven by compact frequency descriptors instead of transmitted camera frames. \nfjf recovers most of the accuracy gain of supervised adaptation across six backbones and three tasks: it closes $82\%$ of the oracle adaptation gap on VWW, $93\% - 95\%$ on ImageNet, and $94\% - 98\%$ on MS-COCO detection (Fig.~\ref{fig:bars_per_arch}, \tabref{tab:taskgen}), while the per-frame uplink never exceeds $832$\,B. In contrast, the entropy minimization (\textit{Oracle - E}) destabilizes on the self-attention and hybrid ImageNet architectures and on VWW.

\textbf{Outlook.}
There are several directions to extend this work. First, the adaptation loop consumes each test \cdag independently, re-initialized from the pre-trained parameters; extending it to \cdag{}s that drift over time would move the framework to a continual setting. Second, the descriptor resolution trades adaptation accuracy against uplink bandwidth; a controller that dynamically selects this resolution per-shift and available computational resources could lower the average uplink overhead without sacrificing adaptation accuracy on the hardest \cdag{}s. Third, since progressive JPEG delivers the \dct DC coefficient of each $8 \times 8$ block  in the first decode pass~\cite{wallace1992jpeg}, the edge device can compute the lowest-resolution descriptor and transmit it while the frame is decoding, overlapping the host adaptation with the remaining decode. 

\textbf{Limitations.} 
First, the figures and tables use a single classifier seed; Sec.~\ref{sec:class_multiseeds} shows that the standard deviation of the post-adaptation accuracy across classifier seeds falls below the standard deviation across transmitted descriptors at every resolution. Second, the adaptation loop reproduces only corruptions from the modeled ten-operator \cdag family (\tabref{tab:dag_stages}): a shift outside this family (\eg motion blur or fog) is mapped to the most similar modeled \cdag in the descriptor space, thus the host adapts on a different corruption. We expect part of the adaptation gain when similar frequency bands are altered and an accuracy below \textit{No Adapt} otherwise, as for the incorrectly predicted test \cdag{}s at resolution $28$ (Fig.~\ref{fig:per_graph_main_grid}). Filtering unreliable stage predictions and extending the corruption family remain future work. Third, whether the adaptation loop transfers to vision-language models increasingly deployed on smart glasses~\cite{liu2025nvila} is untested. We hope this work sparks interest in adaptation systems that communicate compact frequency descriptors rather than camera frames, narrowing the gap between bandwidth-efficient and continuous model adaptation.

\begin{acks}
This research was funded in whole, or in part, by the Austrian Science Fund (FWF) [10.55776/DFH5]. For the purpose of open access, the author has applied a CC BY public copyright license to any Author Accepted Manuscript version arising from this submission. The computational results have been achieved using the Austrian Scientific Computing (ASC) infrastructure.
\end{acks}

\bibliographystyle{ACM-Reference-Format}
\bibliography{references}

\end{document}